%% file: ActiveInference.tex
\documentclass[conference,10pt]{IEEEtran}
\IEEEoverridecommandlockouts

\usepackage{amsmath,amsthm, amssymb,amsfonts,bbm}
\usepackage{cite,cleveref}
\usepackage{algorithm,algorithmic}
\usepackage{enumitem,xcolor}

\usepackage{graphicx}
\usepackage{caption} 
\usepackage{subcaption}	
\graphicspath{{Figures/}}

\newcommand{\upi}{\pi_{\mathrm{upper}}}

\crefname{figure}{fig.}{figs.}
\newcommand{\change}[1]{{#1}}

\usepackage{etoolbox}
\makeatletter 
\pretocmd\@bibitem{\color{black}\csname keycolor#1\endcsname}{}{\fail}
\newcommand\citedcolor[1]{\@namedef{keycolor#1}{\color{blue}}}
\makeatother

\input{Supporting_Files/My_notations}

\begin{document}

\title{Anomaly Detection via Controlled Sensing and Deep Active Inference\\
\thanks{The information, data, or work presented herein was funded in part by National Science Foundation (NSF) under Grant 1618615, Grant 1739748, Grant 1816732 and by the Advanced Research Projects Agency-Energy (ARPA-E), U.S. Department of Energy, under Award Number DE-AR0000940. The views and opinions of authors expressed herein do not necessarily state or reflect those of the United States Government or any agency thereof.}
}
\author{\IEEEauthorblockN{Geethu Joseph, Chen Zhong, M. Cenk Gursoy, Senem Velipasalar, and Pramod K. Varshney}
\IEEEauthorblockA{\textit{Department of of Electrical Engineering and Computer Science } \\
\textit{Syracuse University}\\
 New York 13244, USA}
Emails:\{gjoseph,czhong03,mcgursoy,svelipas,varshney\}@syr.edu.}

\maketitle

\begin{abstract}
In this paper, we address the anomaly detection problem where the objective is to find the anomalous processes among a given set of processes. To this end, the decision-making agent probes a subset of processes at every time instant and obtains a potentially erroneous estimate of the binary variable which indicates whether or not the corresponding process is anomalous. The agent continues to probe the processes until it obtains a sufficient number of measurements to reliably identify the anomalous processes. In this context, we develop a sequential selection algorithm that decides which processes to be probed at every instant to detect the anomalies with an accuracy exceeding a desired value while minimizing the delay in making the decision and the total number of measurements taken. Our algorithm is based on \emph{active inference} which is a general framework to make sequential decisions in order to maximize the notion of \emph{free energy}. We define the free energy using the objectives of the selection policy and implement the active inference framework using a deep neural network approximation. Using numerical experiments, we compare our algorithm with the state-of-the-art method based on deep actor-critic reinforcement learning and demonstrate the superior performance of our algorithm.

\end{abstract}

\section{Introduction}
In many practical applications such as remote health monitoring using sensors, the goal is to identify the anomalies among a given set of functionalities of a system~\cite{chung2006remote,bujnowski2013enhanced}. Here, the system is equipped with multiple sensors and each sensor monitors a different, but not necessarily independent functionality (which we henceforth refer to as a process) of the system. The sensor sends its observations to the decision-making agent over a communication link, and the received observation may be distorted due to the unreliability in the sensor hardware and/or the noisy link (e.g., a wireless channel) between the sensor and the agent. Hence, the decision agent needs to probe each process multiple times before it declares one or more of the processes to be anomalous with the desired confidence. Repeatedly probing all the processes allows the agent to quickly find any potential system malfunction, but this incurs a large cost (e.g., \change{ higher energy consumption} that reduces the life span of the sensor network). Therefore, the agent uses the \emph{controlled sensing} technique with which it probes a small subset of processes at every time instant. In this context, we address the question of how the agent sequentially chooses a subset of processes so that it accurately detects the anomalies with a minimum delay and a minimum number of sensor measurements.

A classical approach to solve the sequential sensor selection problem is based on the active hypothesis testing framework~\cite{zhong2019deep,joseph2020anomaly} where the decision-making agent constructs a hypothesis corresponding to each of the possible states of the processes and determine which one of these hypotheses is true. Active hypothesis testing is a well-studied problem and several solution strategies have been proposed in the literature~\cite{chernoff1959sequential,bessler1960theory,nitinawarat2013controlled,naghshvar2013active,huang2018active}. However, these approaches provide model-based algorithms which are designed under simplified modeling assumptions. This has motivated the researchers to design data-driven deep learning algorithms~\cite{kartik2018policy,zhong2019deep,joseph2020anomaly}. These algorithms are not only more flexible than traditional algorithms, but they also possess reduced computational complexity. The existing literature along these lines relies on the most fundamental reinforcement learning (RL) algorithms such as Q-learning~\cite{kartik2018policy} and actor-critic~\cite{zhong2019deep,joseph2020anomaly}. However, recently a new framework called \emph{active inference} has been shown to be a promising complement to the traditional RL approaches for several sequential decision-making problems~\cite{friston2015active,friston2017active,friston2017active_b}. Therefore, in this paper, we develop and implement a novel policy to select processes to obtain measurements \change{at each step}, inspired by the active inference approach.


The contributions of the paper are as follows: we first define the notion of \emph{free-energy} based on the entropy associated with the estimate of the states of the processes and the cost of sensing. This allows us to reformulate the anomaly detection problem as an active inference problem in which the goal is to minimize the free energy. We then implement our algorithm using deep neural networks which are relatively less explored in the context of active inference. Our algorithm balances the model-based and the data-driven approaches of active inference. Specifically, we use the model-based posterior updates to tackle the uncertainties in the observations, and the data-driven neural network to handle the underlying statistical dependence between the processes. The active inference approach has many similarities to the reinforcement-based algorithms, such as learning probabilistic models, exploration and exploitation of various actions, and efficient planning. \change{So we compare our algorithm with the existing RL-based approach presented  in \cite{joseph2020anomaly} using numerical simulations. We observe that the delay in estimation is smaller for our method while the corresponding accuracy and cost of sensing are competitive  to the performance of the RL-based method given in \cite{joseph2020anomaly}}. This advantage makes our active inference-based approach a better alternative to the existing RL-based method.


\section{Anomaly Detection Problem}\label{sec:anomaly}
We consider $N$ random processes that are potentially statistically dependent. Each process is in one of the two states: normal (denoted by 0) or anomalous (denoted by 1). The states of these processes are denoted by a random vector $\vecs\in\{0,1\}^N$. The goal of the work is to detect the anomalous processes out of the $N$ processes, which is equivalent to estimating the random vector $\vecs$.  The dependence pattern and the number of anomalous processes are unknown to the decision-making agent.

To estimate $\vecs$, the decision-making agent probes one or more processes at every time instant and obtains potentially erroneous observations of the corresponding entries of $\vecs$. Let the set of processes probed at time $k$ be $\calA_k\in\calP$ and the corresponding observation vector be $\vecy_{\calA_k}(k)\in\{0,1\}^{\lv\calA_k\rv}$. Here, $\calP$ denotes the power set of $\{1,2,\ldots,N\}$ without the null set ($\lv\calP\rv=2^{N}-1$). \change{The observation corresponding to the $i\nth$ process at time $k$, denoted by $\vecy_i(k)\in\{0,1\}$, obeys the following probabilistic model:}
\begin{equation}\label{eq:mesurement}
\vecy_{i}(k) = \begin{cases}
\vecs_{i} & {\text{ with probability } 1-p}\\
1-\vecs_{i}& {\text{ with probability } p},
\end{cases}
\end{equation}
where $p\in[0,1]$ denotes the probability that the observation differs from the actual state of the process. We assume that given $\vecs$, the observations obtained across different time instants are jointly (conditionally) independent. Also, probing each process incurs a cost of sensing of $\lambda\geq0$, i.e., the cost of sensing at time $k$ is $\lv\calA_k\rv\lambda$. 

At each time $k$, the agent determines which processes to observe ($\calA_k$)  until it declares the estimate of $\vecs$ with the desired confidence. \change{The selection policy is designed such that the stopping time $K$ and the total cost of sensing $\lambda\sum_{k=1}^K\lv\calA_k\rv$ are minimized}.  

\section{Anomaly Detection Using Deep Active Inference}\label{sec:active_infer}
The active inference framework relies on  a normative theory of brain function based on its perception of the environment. At a high level, the active inference agent maintains a generative model that represents its perception. The generative model $Q$ comprises a joint probability distribution on the state of the environment, the actions, and the corresponding observations. The generative model assigns higher probabilities to the states and actions that are favorable to the agent, and therefore, it is biased towards the agent's preferences. Given a generative model, the agent inverts the model using the method of approximate Bayesian inference. To this end, it defines a variational distribution $q$  that the agent controls. The distribution $q$ is optimized by minimizing the Kullback-Leibler (KL) divergence between the distributions $q$ and $Q$. Therefore, if we choose actions from the distribution $q$, they fulfill the agent's preferences. The KL divergence between the variational distribution and the generative model is called the variational free energy. In short, the goal of the active inference agent is to minimize its expected free energy (EFE) into the future up to the stopping time $K$. Next, we provide the details of the active inference framework in the context of anomaly detection.

\subsection{Environment}
The environment of the active inference framework refers to the set of states, actions, and observations. In the context of our anomaly detection problem, we define the state of the active inference framework at time $k$ as the posterior belief $\pi(k)$ on the random vector $\vecs\in\{0,1\}^N$. Since there are $m=2^N$ possible values for $\vecs$, the posterior belief is an $m-$dimensional vector $\pi\in[0,1]^m$. Further, the actions refer to the selection of which processes to observe $\calA_k\in\calP$, and $\vecy_{\calA_k}$ denotes the observations.

We first note that at time $k$, the information available to the agent is the set of processes observed till time $k$ and the corresponding observation vectors: \change{$\lc\calA_j,\vecy_{\calA_j}\rc_{j=1}^k$}. Using this information, the posterior belief vector $\pi(k)\in[0,1]^m$ can be computed in closed form as follows~\cite{joseph2020anomaly}:
\begin{equation}\label{eq:posterior_update}
\pi_i(k) =  \frac{\pi_i(k-1) \prod_{a\in\calA_k}\ls (1-p) \mathbbm{1}_{\calE_{a,k,i}}+ p\mathbbm{1}_{\calE_{a,k,i}^c} \rs}{\sum_{i=1}^m \pi_i(k-1) \prod_{a\in\calA_k} \ls (1-p) \mathbbm{1}_{\calE_{a,k,i}}+ p\mathbbm{1}_{\calE_{a,k,i}^c}\rs},
\end{equation}
where $\mathbbm{1}$ is the indicator function and the event \change{$\calE_{a,k,i}\triangleq \lc\vecy_a(k)=\vecs_a\middle|\calH=i\rc$} denotes the event that the observation obtained and the corresponding state are the same, when the index  corresponding to the true value of $\vecs$ is $\calH=i$. Also, the event $\calE_{a,k,i}^c\triangleq \lc\vecy_a(k)\neq\vecs_a\middle|\calH=i\rc$ denotes the complement of $\calE_{a,k,i}$. As a result, given the previous state $\pi(k-1)$,  the action $\calA_k$ and the observation $\vecy_{\calA}$, we can exactly compute the updated posterior belief $\pi(k)$ using \eqref{eq:posterior_update}. Therefore, the generative model that learns the environment is a distribution on the actions and the observations: $Q(\calA_{k},\vecy_{\calA_k}|\pi(k-1))$ . 

\subsection{Preferences}
In this subsection, we consider the preferences of the agent that defines the generative model. Recall that our goal is to estimate the vector $\vecs$ with confidence exceeding a specific level while minimizing the stopping time $K$ and the cost of sensing $\lambda\sum_{k=1}^K\lv\calA_k\rv$.  Clearly, the best estimate of $\vecs$ based on the posterior belief corresponds to $i^*(k)\triangleq\underset{i=1,2,\ldots,m}{\arg\max}\pi_i(k)$, and the confidence associated with the estimation is $\pi_{i^*(k)}(k)$. Therefore, the agent terminates the detection algorithm when
\begin{equation}\label{eq:termination}
\underset{i=1,2,\ldots,m}{\arg\max}\pi_i(k)>\upi,
\end{equation}
where $\upi$ is the desired level of confidence.  In short, the decision making relies only on the posterior belief $\pi(k)$. Also, as $k$ increases, we get more observations and the posterior belief becomes more accurate. Therefore, the selection policy $\mu$ is a function of the latest value of the posterior belief: $\mu(\pi(k-1))=\calA_k$. 

Further exploring the objective of the policy design, we note that minimizing the stopping time is identical to driving the largest entry of $\pi(k)$ to $\upi$ as soon as possible. We achieve this by minimizing the entropy $H(\pi(K))$ of $\pi(K)$ because the entropy is minimized when the largest entry of $\pi(K)$ is 1 and the remaining entries are zeros. Here, the entropy is given by
\begin{equation}
H(\pi) =-\sum_{i=1}^m\pi_i\log(\pi_i). 
\end{equation}
We note that this approach is different from the Bayesian log likelihood ratio based-approach in \cite{kartik2018policy,zhong2019deep,joseph2020anomaly}. 
Therefore, we define the instantaneous objective function that the agent aims to minimize at time $k$ as follows:
\begin{equation}\label{eq:reward}
r(k) =  H(\pi(k))-H(\pi(k-1))+\lambda\lv\calA_k\rv.
\end{equation}
This definition ensures that the overall objective function is given by
\begin{equation}\label{eq:total_reward}
\sum_{k=1}^K r(k) =  H(\pi(K))-H(\pi(0))+\sum_{k=1}^K\lambda\lv\calA_k\rv,
\end{equation}
where minimizing $H(\pi(K))-H(\pi(0))$ minimizes the entropy in the posterior belief as $H(\pi(0))$ is a constant, and minimizing $\sum_{k=1}^K\lambda\lv\calA_k\rv$ minimizes the total cost of sensing. The instantaneous objective function $r(k)$ represents the preferences of the agent at time $k$ and it is encoded into the generative model as the prior probability on the belief vector:
\begin{multline}\label{eq:observation_prob}
Q(\vecy_{\calA_k}|\calA_{k},\pi(k-1)) \\= \sigma\lb \change{-H(\pi(k))}+H(\pi(k-1))-\lambda\lv\calA_k\rv\rb,
\end{multline}
where $\sigma(\cdot)$ is the softmax function. Also, $\pi(k)$ is a function of $\pi(k-1),\calA_k$ and $\vecy_{\calA_k}$ due to \eqref{eq:posterior_update}. We also note that 
\begin{equation}
Q(\vecy_{\calA_k},\calA_{k}|\pi(k-1)) = Q(\vecy_{\calA_k}|\calA_{k},\pi(k-1))Q(\calA_k|\pi(k-1)).
\end{equation}
Therefore, the generative model is completely defined if we specify the distribution $Q(\calA_k|\pi(k-1))$. This distribution is defined based on the EFE of the future as we discuss in the following subsection.

\subsection{Total expected free energy}
The variational free energy $F$ is the KL divergence between the variational distribution $q(\calA|\pi(k-1))$ and the generative model $Q(\calA|\pi(k-1))$. Thus,
\begin{equation}\label{eq:F_defn}
F(k) = \sum_{\calA\in\calP}q(\calA|\pi(k-1))\log\frac{q(\calA|\pi(k-1))}{Q(\calA|\pi(k-1))}.
\end{equation}
The goal of the agent is to minimize the total free-energy of the expected trajectories into the future:
\begin{equation}\label{eq:G_defn}
G(\calA,\pi) =  \sum_{j=k}^K \expect{}{F(j)\middle|\calA_k=\calA,\pi(k-1)=\pi}.
\end{equation}
In other words, the agent computes the expected free-energy of all paths into the future and probabilistically chooses an action that minimizes the expected free-energy. Therefore, a popular choice for  the distribution over the actions assigned by the generative model is a Boltzmann distribution over the expected free energies~\cite{friston2015active, schwartenbeck2019computational,millidge2020deep}:
\begin{equation}\label{eq:action_distribution}
Q(\calA|\pi(k-1)) = \sigma\lb-G(\calA,\pi(k-1))\rb,
\end{equation}
where $\sigma(\cdot)$ is again the softmax function, and $G$ is given by \eqref{eq:G_defn}. 

So far, we have presented the conceptual aspects of our algorithm. We next discuss how to compute the expressions in \eqref{eq:F_defn} and \eqref{eq:G_defn}.

\subsection{Deep-learning based implementation}
We implement our algorithm using deep neural networks. We start with the computation of the free energy in \eqref{eq:F_defn}: 
\begin{multline}
F = - H(q(\calA|\pi(k-1))) \\
- \sum_{\calA\in\calP}q(\calA|\pi(k-1))\log Q(\calA|\pi(k-1)),\label{eq:F_update}
\end{multline}
where the entropy term \change{$H(q(\mathcal{A}|\pi(k-1)))$} is a function of the variational distribution $q$ which is controlled by the agent. We implement this distribution using a neural network which we refer to as the \emph{policy network}. The policy neural network takes the posterior belief $\pi(k-1)$ as the input and outputs  stochastic selection policy $q_{\theta}\in[0,1]^{m-1}$ which is a probability distribution on $\calP$ and parameterized by $\theta$. Therefore, the entropy term is computed using the entropy of the distribution outputted by the neural network. This neural network also gives the policy implemented by the agent, which is sampled from the distribution $q$ learned at time $k$:
\begin{equation}
\calA_k=\mu(\pi(k-1)) \sim q_{\theta}(\pi(k-1)).
\end{equation}

Further, the second term in \eqref{eq:F_update} can be determined using \eqref{eq:action_distribution} and \eqref{eq:G_defn}. From \eqref{eq:G_defn}, the EFE for a single time-step can be approximated as follows~\cite{millidge2020deep}:
\begin{multline}
G(\calA_k,\pi(k-1)) \approx \change{-\log Q(\vecy_{\calA_k}|\calA_{k},\pi(k-1))} \\ +   \expect{\calA\sim Q(\cdot|\pi(k)}{G(\calA,\pi(k))}.\label{eq:G_expand}
\end{multline}
Here, the first term is determined using \eqref{eq:observation_prob}. However, the second term in \eqref{eq:G_expand} involves explicit computation into the future values. Therefore, we learn a bootstrap estimate of this quantity using a neural network which we refer to as the \emph{bootstrapped EFE-network}. Let $G_{\phi}(\calA)$ denote this neural network where $\phi$ is the parameter of the network. In other words, the estimate of the neural network is the predicted value of the free-energy of the system. Thus, \eqref{eq:G_expand} reduces to
\begin{multline}\label{eq:bootstrap}
G(\calA_k,\pi(k-1)) = H(\pi(k))-H(\pi(k-1))\\+\lambda\lv\calA_k\rv+\expect{\calA\sim Q(\cdot|\pi(k)}{ G_{\phi}(\calA_{k+1},\pi(k))}.
\end{multline}
Substituting \eqref{eq:bootstrap} and \eqref{eq:action_distribution} into \eqref{eq:F_update} completes the derivation of the algorithm. 

To summarize, our solution involves two neural networks $q_{\theta}$ and $G_{\phi}$ which represent the policy and the expected free-energy, respectively. At every time instant, we sample an action from the output distribution of the policy network $q_{\theta}$ and obtain the corresponding observation $\vecy_{\calA_k}$. Next, we compute the bootstrapped EFE estimate and the variational free energy using the neural networks and \eqref{eq:F_update} and \eqref{eq:bootstrap}. Finally, the parameter $\theta$ of the policy network is modified by minimizing the variational free energy $F(k)$. Similarly, the parameter $\phi$ of the bootstrapped EFE-network is optimized by comparing EFE-network output with the value of the expected value $G(\calA)$ calculated at time $k$. We use the $\ell_2-$norm of the difference between the two estimates:
\begin{equation}\label{eq:bootstraploss}
L = \lV G_{\phi}(\calA)-G(\calA)\rV^2.
\end{equation}
The pseudo-code of the algorithm is summarized in \Cref{alg:ActiveInference} below.

\begin{algorithm}[ht]
\caption{Active inference for anomaly detection}
\label{alg:ActiveInference}
\begin{algorithmic}[1]
\ENSURE 
\begin{itemize}[leftmargin=0.3cm]
\item Policy network $q_{\theta}(a|\pi)$ with parameters $\theta$
\item Bootstrapped EFE-network $G_{\phi} (\pi; a)$ with parameters $\phi$  
\end{itemize}
\REPEAT
\STATE Initialize the prior state $\pi_0\in[0,1]^m$ (can be  learned from the training data)
\STATE Time index $k =0$
\WHILE {$k < T$ and $\underset{i}{\max} \; \pi_i > \upi$ and $k<T_{\max}$} 
\STATE Choose action $\calA_k\sim q_{\theta}(\pi(k-1))$
\STATE Generate observations $\vecy_{\calA_k,k}$
\STATE Compute $\pi(k+1)$ using \eqref{eq:posterior_update}
\STATE Compute the bootstrapped EFE estimate $G$ using \eqref{eq:bootstrap}
\STATE Compute the variational free energy $F$ using \eqref{eq:action_distribution} and \eqref{eq:F_update}
\STATE Update the policy network network by minimizing the variational free energy $F$ with respect to $\theta$
\STATE Update the bootstrapped EFE-network by minimizing the boostrapping loss in \eqref{eq:bootstraploss} with respect to $\phi$
\STATE Increase time index $k=k+1$
\ENDWHILE
\UNTIL
\STATE Declare the estimate corresponding to $\underset{i}{\arg\max} \; \pi_i$
\end{algorithmic}
\end{algorithm}

\section{Numerical Results}\label{sec:simulations}
In this section, we present numerical results comparing our algorithm with the actor-critic method in \cite{joseph2020anomaly}. The simulation setup is similar to that in \cite{joseph2020anomaly}. We choose  the number of processes as $N=3$ and thus, $m=2^N=8$.  The probability of a process being normal is taken as $q=0.8$. Here, the first and second processes are assumed to be statistically dependent, and the third process is independent of the other two.  The correlation between the dependent processes is captured by the parameter $\rho\in[0,1]$:
\begin{align}
\bbP\lc \vecs_1=0,\vecs_2=0\rc &= q^2+\rho q(1-q)\\
\bbP\lc \vecs_1=0,\vecs_2=1\rc &= q(1-q)(1-\rho)\\
\bbP\lc \vecs_1=1,\vecs_2=0\rc &= q(1-q)(1-\rho)\\
\bbP\lc \vecs_1=1,\vecs_2=1\rc &= (1-q)^2+\rho q(1-q).
\end{align}
Also, we assume that the crossover probability of the observations is $p=0.8$, and the maximum number of time slots for each episode (trial or run) is $T_{\max}=300$.

For the active inference algorithm, we implement the policy neural network and the bootstrapped EFE-network with three layers and the ReLU activation function between \change{consecutive layers}. To update the parameters of the neural networks, we apply the Adam Optimizer, and we set the learning rates of the policy network and the bootstrapped EFE-network as $10^{-6}$ and $5\times 10^{-6}$, respectively. The implementation of the actor-critic method is the same as that in \cite{joseph2020anomaly} except that we use the entropy based-reward function as defined in \eqref{eq:reward}. Also, we choose the learning rates of the actor and critic networks as $5\times 10^{-4}$ and $5\times 10^{-3}$, respectively.

\begin{figure*}[hp]
\begin{center}
	\begin{subfigure}[b]{5.7cm}
		{\includegraphics[width=5.5cm]{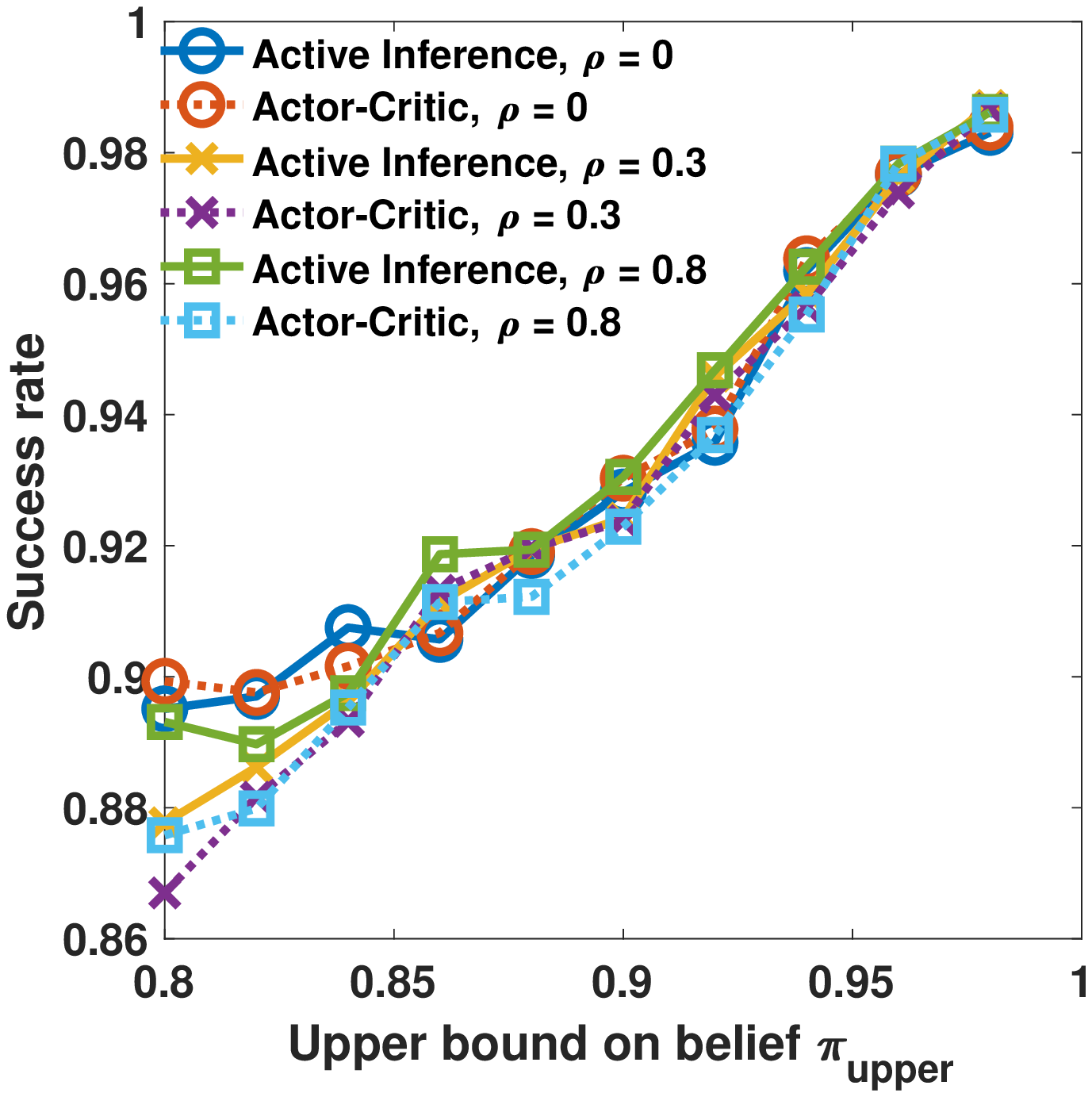}}
		\caption{Cost per measurement $\lambda=0.05$}
	 \end{subfigure}
	 \hspace{0.3cm}
     \begin{subfigure}[b]{5.7cm}
		{\includegraphics[width=5.5cm]{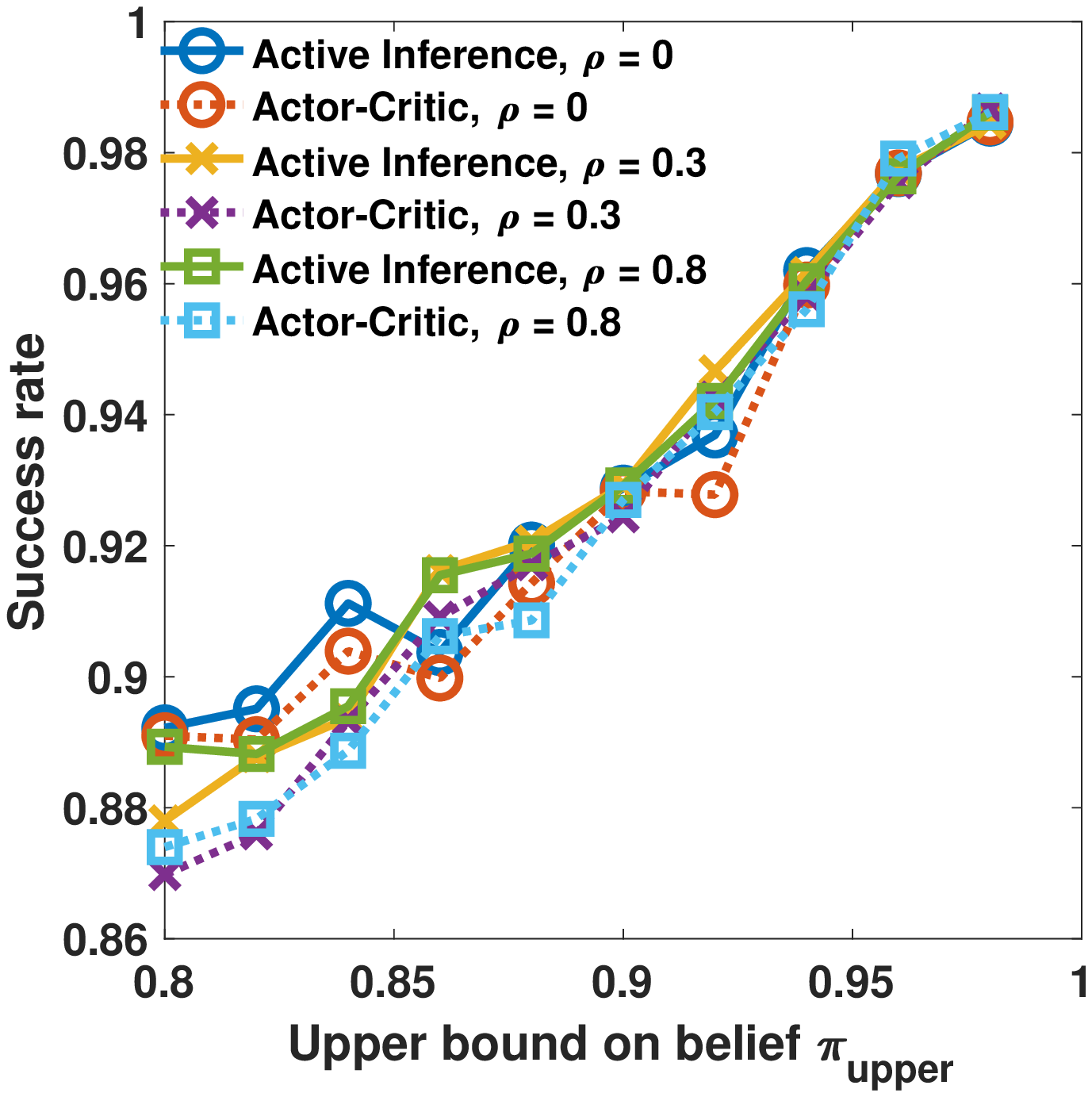}}	
		\caption{Cost per measurement $\lambda=0.1$}	
      \end{subfigure}
      \hspace{0.3cm}
	\begin{subfigure}[b]{5.7cm}
		{\includegraphics[width=5.5cm]{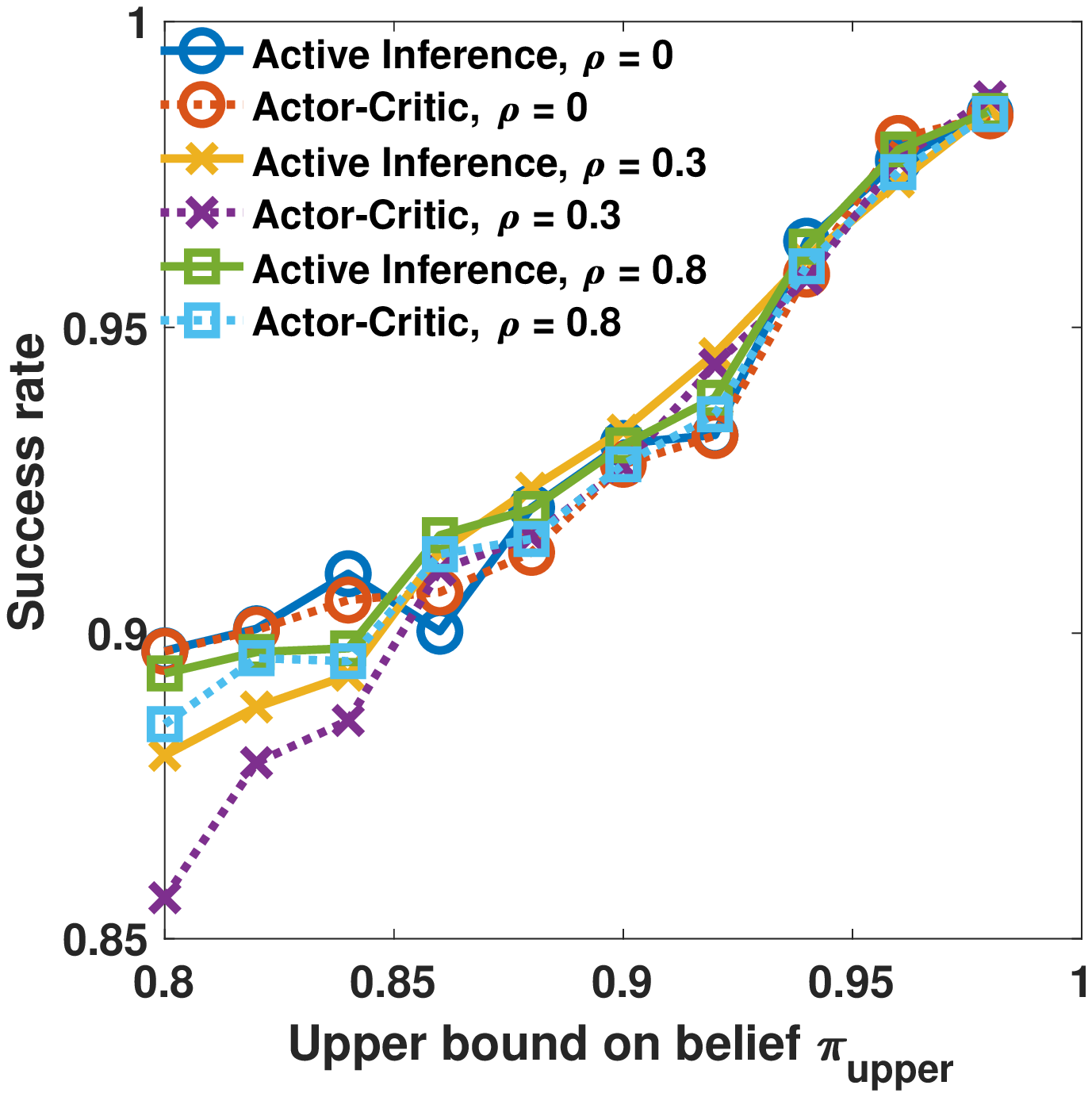}}
		\caption{Cost per measurement $\lambda=0.2$}	
	\end{subfigure}
\end{center}
		\caption{Variation of the success rate of the active inference and the actor-critic algorithms when $\upi,\lambda$ and $\rho$ are varied.}
	\label{fig:accuracy_Acc}
\end{figure*}

\begin{figure*}[hp]
\begin{center}
		\begin{subfigure}[b]{5.7cm}
		   {\includegraphics[width=5.5cm]{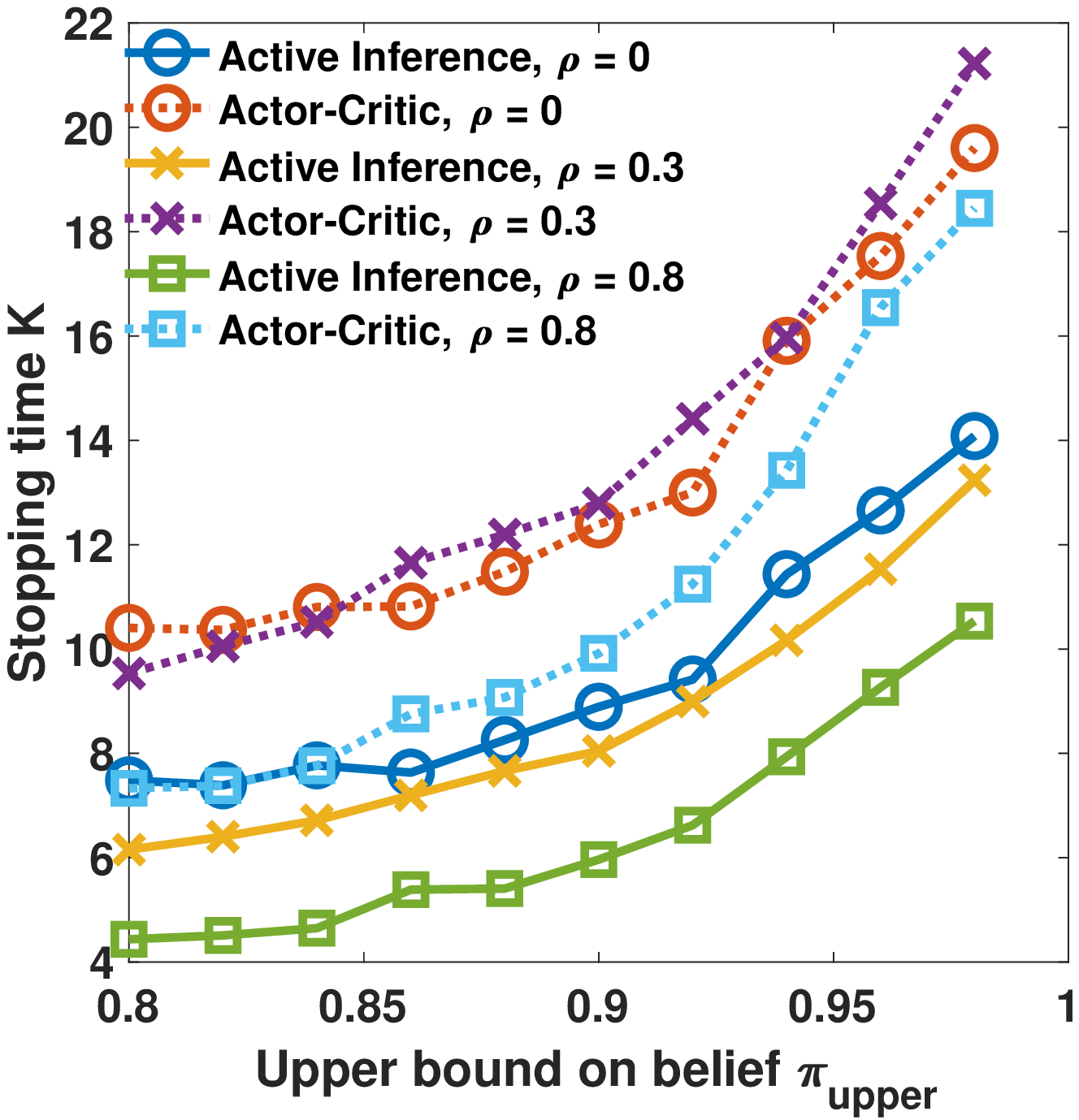}}
			\caption{Cost per measurement $\lambda=0.05$}
		\end{subfigure}
			 \hspace{0.3cm}
		\begin{subfigure}[b]{5.7cm}
		     {\includegraphics[width=5.5cm]{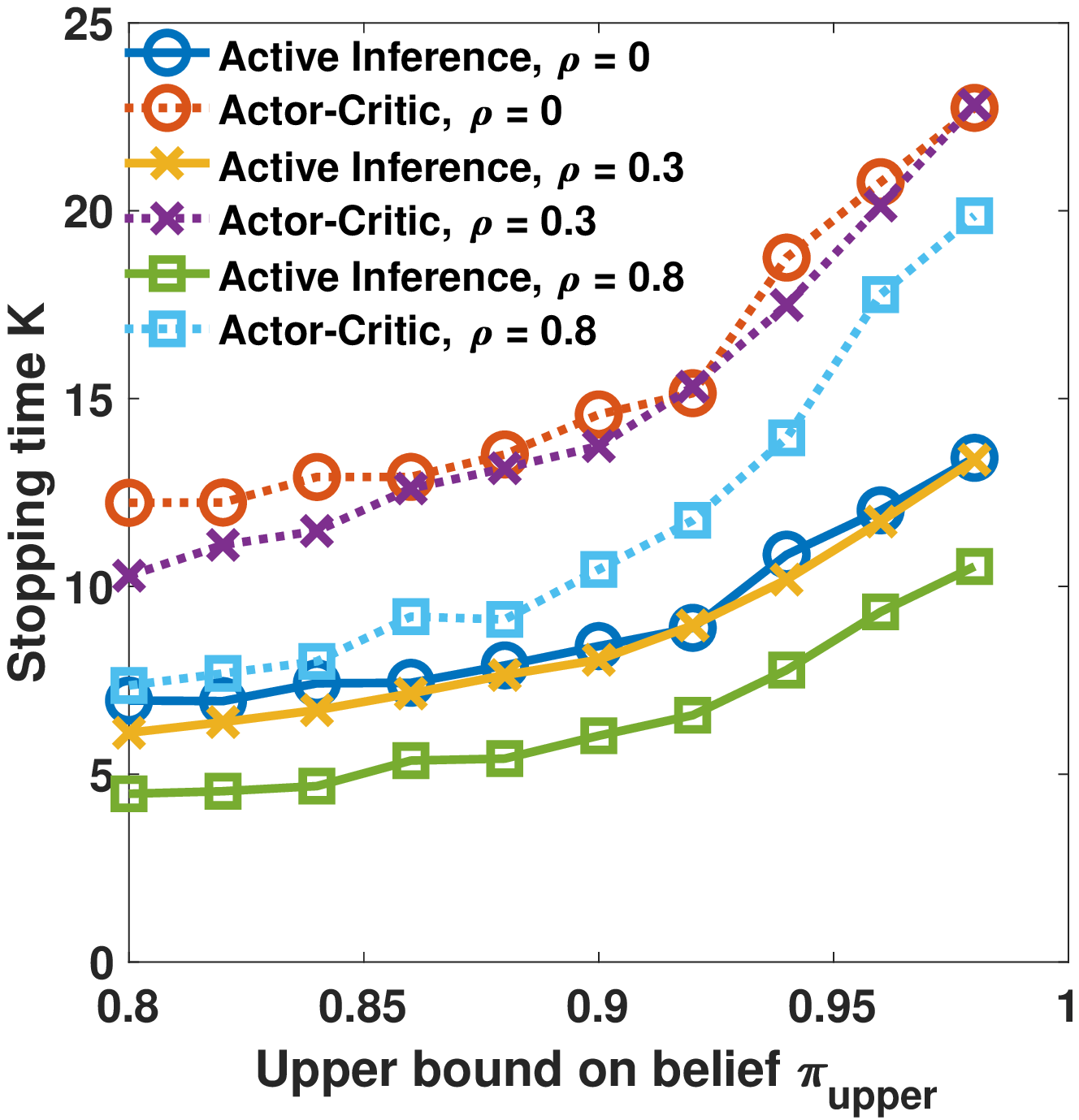}}
			\caption{Cost per measurement $\lambda=0.1$}	
		\end{subfigure}
			 \hspace{0.3cm}
		\begin{subfigure}[b]{5.7cm}
		      {\includegraphics[width=5.5cm]{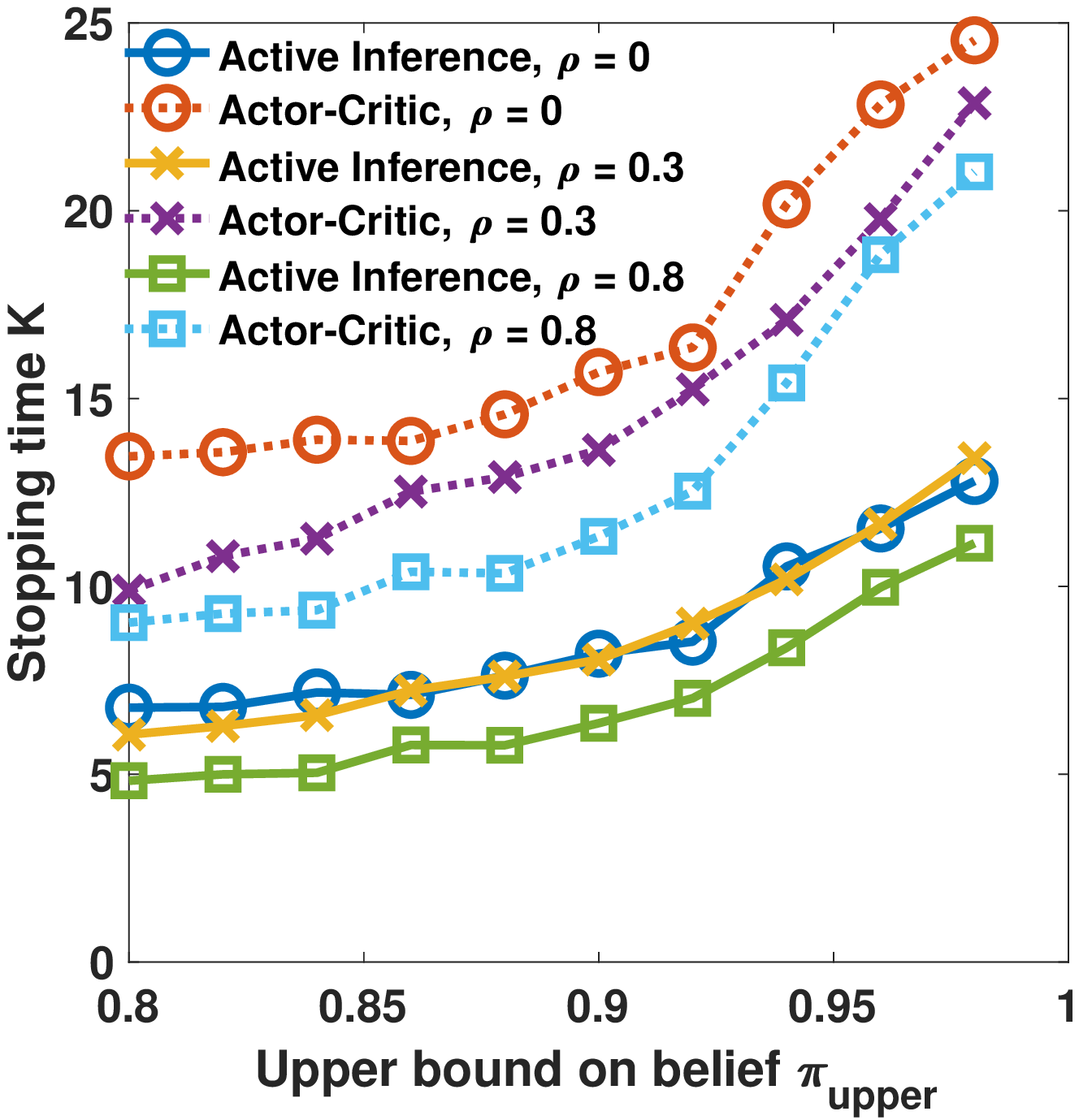}}
			 \caption{Cost per measurement $\lambda=0.2$}
		\end{subfigure}		
\end{center}
	\caption{Variation of the stopping time $K$ of the active inference and the actor-critic algorithms when $\upi,\lambda$ and $\rho$ are varied.}
	\label{fig:delay_Acc}
\end{figure*}

\begin{figure*}[hp]
\begin{center}
		\begin{subfigure}[b]{5.7cm}
		{\includegraphics[width=5.5cm]{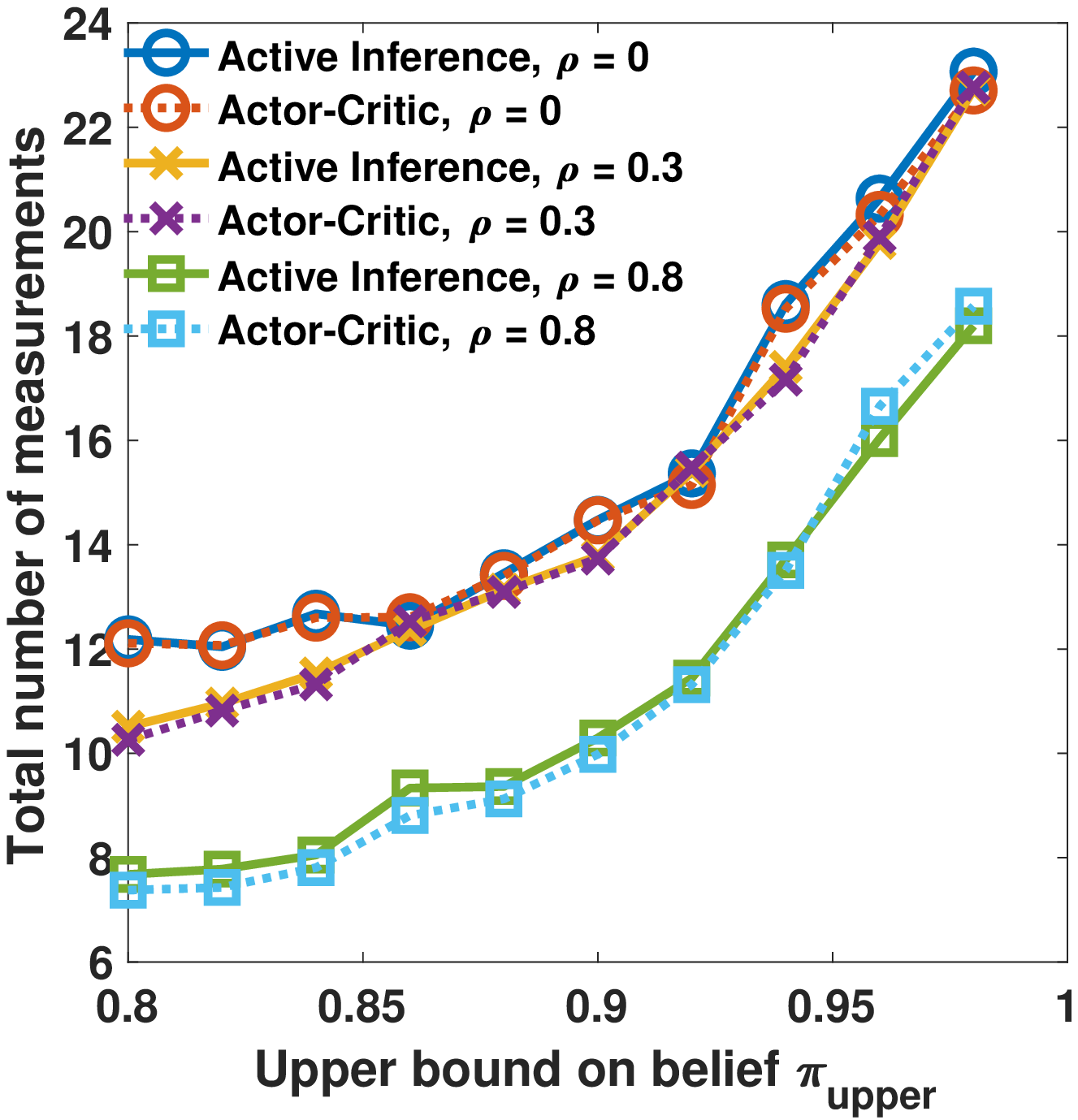}}
			\caption{Cost per measurement $\lambda=0.05$}
			\end{subfigure}
				 \hspace{0.3cm}
		\begin{subfigure}[b]{5.7cm}
			{\includegraphics[width=5.5cm]{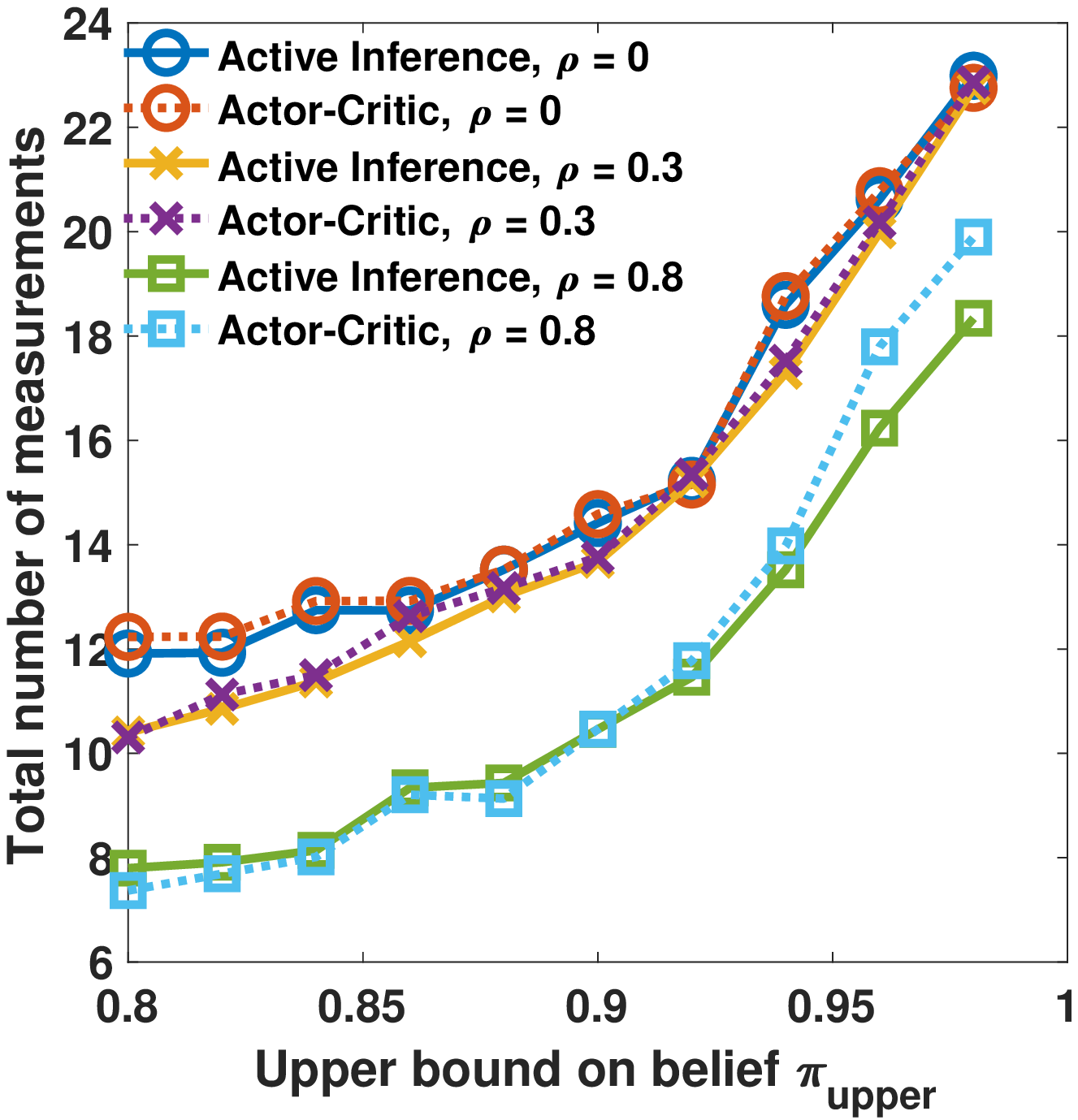}}
			\caption{Cost per measurement $\lambda=0.1$}	
		\end{subfigure}
			 \hspace{0.3cm}
		\begin{subfigure}[b]{5.7cm}
			{\includegraphics[width=5.5cm]{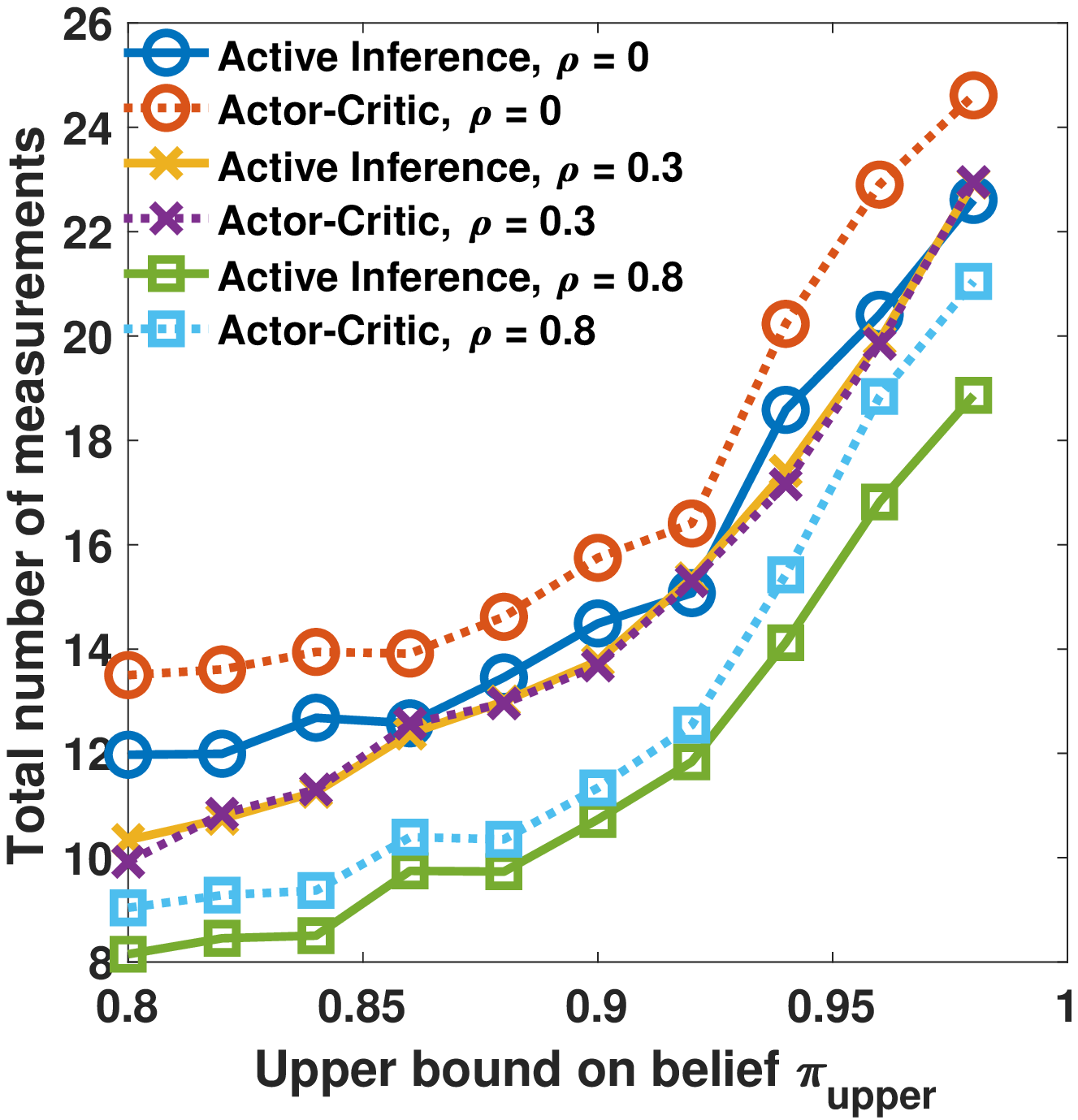}}
			\caption{Cost per measurement $\lambda=0.2$}
		\end{subfigure}		
\end{center}
	\caption{ Variation of the total number of measurements $\sum_{k=1}^K\lv\calA_k\rv$ of the active inference and the actor-critic algorithms when $\upi,\lambda$ and $\rho$ are varied.}
	\label{fig:totalcost}
\end{figure*}


The simulation results are presented in \Cref{fig:accuracy_Acc,fig:delay_Acc,fig:totalcost}. Our observations from the numerical results are as follows:
\begin{itemize}
\item \emph{Success rate:} In \Cref{fig:accuracy_Acc}, we plot the success rates of the two algorithms as a function of the upper bound on the posterior  $\upi$. The success rate is defined as the ratio between the number of times the algorithm correctly identifies all the anomalous processes to the total number of trials. We observe that the success rates achieved by both algorithms are comparable in all the settings. Also, the success rate depends \change{primarily} on $\upi$ and it \change{is almost} insensitive to $\lambda$ and $\rho$. This is intuitive because $\upi$ sets the confidence level with which the algorithms identify the anomalies, and therefore, for the same confidence level, the success rates achieved by the algorithms are almost the same. 

\item \emph{Stopping time:} In \Cref{fig:delay_Acc}, we show the variation of the stopping time $K$ with $\upi$. We see that the stopping time increases with $\upi$ in all cases, as a higher value of $\upi$ requires the algorithms to collect more observations before they make the decision regarding the anomalous processes. Also, we observe that the stopping time decreases with an increase in $\rho$ for all values of $\lambda$ and $\upi$. This decrease is expected due to the fact that as the correlation increases, an observation corresponding to one of the dependent processes gives more information about the other. Consequently, the algorithms require \change{fewer observations}, and thus, a smaller stopping time, to achieve  the same confidence level. \change{Finally, we notice that the stopping time for the active inference algorithm is less than that of the actor-critic algorithm.}

\item \emph{Total number of measurements: } \Cref{fig:totalcost} compares the total number of measurements $\sum_{k=1}^K\lv\calA_k\rv$ obtained by the two algorithms in different settings. Clearly, the total number of measurements decreases with $\rho$, which is expected as mentioned above. Also, we infer that the total number of measurements obtained by both algorithms are similar in all the settings with the active inference algorithm collecting slightly fewer measurements compared to the actor-critic algorithm. 

\end{itemize}

Thus, we conclude that the two algorithms achieve comparable success rates and incur a similar total cost of sensing, but the active inference algorithm has better stopping time compared to the actor-critic algorithm. This indicates that our algorithm identifies the anomalies faster than the actor-critic algorithm. Moreover, the stopping time of our algorithm does not vary much with $\lambda$ while the stopping time of the actor-critic algorithm increases with $\lambda$.  This implies that the actor-critic algorithm is more sensitive to the instantaneous cost of sensing $\lambda\lv\calA_k\rv$ than the total cost of sensing $\sum_{k=1}^K\lambda\lv\calA_k\rv$. To elaborate, we note that both algorithms continue to acquire measurements until the desired level confidence level $\upi$ is achieved. However, since the actor-critic algorithm optimizes the average cost of sensing $\frac{1}{K}\sum_{k=1}^K\lambda\lv\calA_k\rv$, as $\lambda$ increases, it picks a fewer number of processes per time instant and this results in an increased stopping time. On the contrary, the average number of processes selected by our algorithm does not vary much with $\lambda$. \change{Therefore, we achieve better performance by carefully designing the objective function using a novel entropy based-function and the total cost of sensing whereas the actor-critic algorithm optimizes the average change in entropy and the average cost of sensing.}


\section{Conclusion}

In this paper, we presented an anomaly detection algorithm using an active inference-based approach. We modeled the problem of anomaly detection as an active inference problem aiming at the detection accuracy exceeding a desired value while minimizing the delay and total cost of sensing. We designed a new objective function based on entropy and implemented the active inference algorithm using a deep learning-based approach. Through simulation results, we compared our algorithm with an algorithm based on the deep actor-critic method in terms of the success rate, stopping time, and total cost of sensing. \change{The results demonstrated that our algorithm can detect the anomalies quicker (as indicated by the smaller stopping times) and achieves a competitive success rate with a similar cost of sensing as the actor-critic algorithm.} However, we detect all the anomalous processes at a given time, assuming that the (normal or anomalous) behaviors of the processes remain unchanged until the agent makes a decision. Extending our algorithm to track any changes in the behavior of the processes over a longer time period is an interesting direction for future work.

\bibliographystyle{IEEEtran}
\bibliography{Supporting_Files/IEEEabrv,Supporting_Files/bibJournalList,Supporting_Files/AnomalyDetection}

\end{document}

%% file: Supporting_Files/My_notations.tex
\crefname{app}{Appendix}{Appendices}
\crefname{cor}{Corollary}{Corollary}
\crefname{prop}{Proposition}{Proposition}
\crefname{lemma}{Lemma}{Lemma}
\crefname{defn}{Definition}{Definition}
\crefname{conj}{Conjecture}{Conjecture}
\crefname{exam}{Example}{Example}
\crefname{supp}{Supplemental Section}{Supplemental Section}

\newcommand{\bs}{\boldsymbol}
\newcommand{\bb}{\mathbb}
\newcommand{\mcal}{\mathcal}

\newcommand{\lb}{\left(}
\newcommand{\rb}{\right)}
\newcommand{\ls}{\left[}
\newcommand{\rs}{\right]}
\newcommand{\lc}{\left\{}
\newcommand{\rc}{\right\}}

\newcommand{\lv}{\left\vert}
\newcommand{\rv}{\right\vert}
\newcommand{\lV}{\left\Vert}
\newcommand{\rV}{\right\Vert}

\newcommand{\LRV}[1]{{\left\vert\kern-0.25ex\left\vert\kern-0.25ex\left\vert #1 \right\vert\kern-0.25ex\right\vert\kern-0.25ex\right\vert}}

\newcommand{\expect}[2]{\bb{E}_{#1}\lc#2\rc}

\newcommand{\nth}{^\mathsf{th}}

\newcommand{\bbP}{\bb{P}}

\newcommand{\calA}{\mcal{A}}

\newcommand{\calE}{\mcal{E}}

\newcommand{\calH}{\mcal{H}}

\newcommand{\calP}{\mcal{P}}

\newcommand{\vecs}{\bs{s}}

\newcommand{\vecy}{\bs{y}}

%% file: ActiveInference.bbl
\begin{thebibliography}{10}
\providecommand{\url}[1]{#1}
\csname url@samestyle\endcsname
\providecommand{\newblock}{\relax}
\providecommand{\bibinfo}[2]{#2}
\providecommand{\BIBentrySTDinterwordspacing}{\spaceskip=0pt\relax}
\providecommand{\BIBentryALTinterwordstretchfactor}{4}
\providecommand{\BIBentryALTinterwordspacing}{\spaceskip=\fontdimen2\font plus
\BIBentryALTinterwordstretchfactor\fontdimen3\font minus
  \fontdimen4\font\relax}
\providecommand{\BIBforeignlanguage}[2]{{%
\expandafter\ifx\csname l@#1\endcsname\relax
\typeout{** WARNING: IEEEtran.bst: No hyphenation pattern has been}%
\typeout{** loaded for the language `#1'. Using the pattern for}%
\typeout{** the default language instead.}%
\else
\language=\csname l@#1\endcsname
\fi
#2}}
\providecommand{\BIBdecl}{\relax}
\BIBdecl

\bibitem{chung2006remote}
W.-Y. Chung and S.-J. Oh, ``Remote monitoring system with wireless sensors
  module for room environment,'' \emph{Sensors Actuators B: Chemical}, vol.
  113, no.~1, pp. 64--70, Jan. 2006.

\bibitem{bujnowski2013enhanced}
A.~Bujnowski, J.~Ruminski, A.~Palinski, and J.~Wtrorek, ``Enhanced remote
  control providing medical functionalities,'' in \emph{Proc. Inter. Conf.
  Pervasive Comput. Tech Healthc. Workshops}, May 2013, pp. 290--293.

\bibitem{zhong2019deep}
C.~Zhong, M.~C. Gursoy, and S.~Velipasalar, ``Deep actor-critic reinforcement
  learning for anomaly detection,'' in \emph{Proc. Globecom}, Dec. 2019.

\bibitem{joseph2020anomaly}
G.~Joseph, M.~C. Gursoy, and P.~K. Varshney, ``Anomaly detection under
  controlled sensing using actor-critic reinforcement learning,'' in
  \emph{Proc. IEEE Inter. Workshop SPAWC}, May 2020.

\bibitem{chernoff1959sequential}
H.~Chernoff, ``Sequential design of experiments,'' \emph{Ann. Math. Stat.},
  vol.~30, no.~3, pp. 755--770, Sep. 1959.

\bibitem{bessler1960theory}
S.~A. Bessler, ``Theory and applications of the sequential design of
  experiments, k-actions and infinitely many experiments: Part {I} - theory,''
  Stanford Univ CA Applied Mathematics and Statistics Labs, Tech. Rep., 1960.

\bibitem{nitinawarat2013controlled}
S.~Nitinawarat, G.~K. Atia, and V.~V. Veeravalli, ``Controlled sensing for
  multihypothesis testing,'' \emph{{IEEE} Trans. Autom. Control}, vol.~58,
  no.~10, pp. 2451--2464, May 2013.

\bibitem{naghshvar2013active}
M.~Naghshvar, T.~Javidi \emph{et~al.}, ``Active sequential hypothesis
  testing,'' \emph{Ann. Stat.}, vol.~41, no.~6, pp. 2703--2738, 2013.

\bibitem{huang2018active}
B.~Huang, K.~Cohen, and Q.~Zhao, ``Active anomaly detection in heterogeneous
  processes,'' \emph{{IEEE} Trans. Inf. Theory}, vol.~65, no.~4, pp.
  2284--2301, Aug. 2018.

\bibitem{kartik2018policy}
D.~Kartik, E.~Sabir, U.~Mitra, and P.~Natarajan, ``Policy design for active
  sequential hypothesis testing using deep learning,'' in \emph{Proc.
  Allerton}, Oct. 2018, pp. 741--748.

\bibitem{friston2015active}
K.~Friston, F.~Rigoli, D.~Ognibene, C.~Mathys, T.~Fitzgerald, and G.~Pezzulo,
  ``Active inference and epistemic value,'' \emph{J. Cogn. Neurosci.}, vol.~6,
  no.~4, pp. 187--214, Oct. 2015.

\bibitem{friston2017active}
K.~Friston, T.~FitzGerald, F.~Rigoli, P.~Schwartenbeck, and G.~Pezzulo,
  ``Active inference: {A} process theory,'' \emph{Neural Comput.}, vol.~29,
  no.~1, pp. 1--49, Jan. 2017.

\bibitem{friston2017active_b}
K.~J. Friston, M.~Lin, C.~D. Frith, G.~Pezzulo, J.~A. Hobson, and S.~Ondobaka,
  ``Active inference, curiosity and insight,'' \emph{Neural Comput.}, vol.~29,
  no.~10, pp. 2633--2683, Oct. 2017.

\bibitem{schwartenbeck2019computational}
P.~Schwartenbeck, J.~Passecker, T.~U. Hauser, T.~H. FitzGerald, M.~Kronbichler,
  and K.~J. Friston, ``Computational mechanisms of curiosity and goal-directed
  exploration,'' \emph{Elife}, vol.~8, p. e41703, 2019.

\bibitem{millidge2020deep}
B.~Millidge, ``Deep active inference as variational policy gradients,''
  \emph{J. Math. Psychol.}, vol.~96, p. 102348, Jan. 2020.

\end{thebibliography}
